\def\year{2022}\relax
\documentclass[letterpaper]{article} %
\usepackage{aaai22}  %
\usepackage{times}  %
\usepackage{helvet}  %
\usepackage{courier}  %
\usepackage[hyphens]{url}  %
\usepackage{graphicx} %
\usepackage{natbib}  %
\usepackage{caption} %
\DeclareCaptionStyle{ruled}{labelfont=normalfont,labelsep=colon,strut=off} %
\usepackage{algorithm}
\usepackage{algorithmic}

\usepackage{newfloat}
\usepackage{listings}
\floatstyle{ruled}
\newfloat{listing}{tb}{lst}{}
\floatname{listing}{Listing}
\usepackage{latexsym}
\usepackage{framed}
\usepackage{subfigure}
\usepackage{soul}
\usepackage{csquotes}

\usepackage{tikz}
\usepackage{pgfplots}

\pgfplotsset{compat=1.17,height=2.3in}

\usepackage{booktabs}

\definecolor{red}{HTML}{E31A1C}
\definecolor{blue}{HTML}{1F78B4}
\definecolor{green}{HTML}{33A02C}
\definecolor{orange}{HTML}{FF7F00}
\definecolor{purple}{HTML}{6A3D9A}

\newcommand{\eg}{\textit{e.g.}}

\newcommand{\speaker}[1]{\textbf{\textsc{#1}}}
\definecolor{lightgrey}{rgb}{.9,.9,.9}
\sethlcolor{lightgrey}

\newcommand{\defn}[1]{\textbf{#1}}
\newcommand{\citepos}[1]{\citeauthor{#1}'s (\citeyear{#1})}
\newcommand{\sdg}[1]{\textbf{SDG \##1}}
\newcommand{\sdgs}[2]{\textbf{SDGs \##1\&\##2}}
\newcommand{\sdgstext}[2]{\textbf{SDGs \##1 \& \##2}}

\newcommand{\notes}[1]{\noindent \textbf{\textit{Notes.}} #1}

\newcommand{\multiwoz}{MultiWOZ}

\title{Conversational AI Systems for Social Good:\\ Opportunities and Challenges}

\author{
     Peng Qi,
     Jing Huang,
     Youzheng Wu,
     Xiaodong He,
     Bowen Zhou
}
\affiliations{
  JD AI Research \\
  Mountain View, CA 94043\\
  {\{peng.qi, jing.huang, wuyouzheng1, xiaodong.he, bowen.zhou\}@jd.com}
}

\usepackage{bibentry}

\begin{document}

\maketitle

\begin{abstract}
Conversational artificial intelligence (ConvAI) systems have attracted much academic and commercial attention recently, making significant progress on both fronts.
However, little existing work discusses how these systems can be developed and deployed for social good in real-world applications, with comprehensive case studies and analyses of pros and cons.
In this paper, we briefly review the progress the community has made towards better ConvAI systems and reflect on how existing technologies can help advance social good initiatives from various angles that are unique for ConvAI, or not yet become common knowledge in the community.
We further discuss about the challenges ahead for ConvAI systems to better help us achieve these goals and highlight the risks involved in their development and deployment in the real world.
\end{abstract}

\section{Introduction}

\begin{figure}
    \centering
     \pgfplotstableread{
    year dialog chat conversation
    1974 1 0 0
1975 1 0 0
1976 0 0 0
1977 1 0 0
1978 5 0 2
1979 1 0 0
1980 3 0 2
1981 1 0 1
1982 4 0 2
1983 2 0 0
1984 1 0 0
1985 3 0 0
1986 5 0 1
1987 3 0 0
1988 11 0 0
1989 4 0 1
1990 13 0 1
1991 9 0 2
1992 15 0 4
1993 10 0 1
1994 14 1 1
1995 10 0 1
1996 9 0 1
1997 37 0 4
1998 37 0 1
1999 19 0 0
2000 58 0 8
2001 37 0 4
2002 53 0 5
2003 54 2 10
2004 78 1 10
2005 51 2 9
2006 72 3 12
2007 73 2 17
2008 73 2 16
2009 82 0 11
2010 84 4 20
2011 57 3 10
2012 93 8 27
2013 67 4 19
2014 73 4 24
2015 62 3 22
2016 84 12 43
2017 94 15 27
2018 143 20 84
2019 176 19 103
2020 276 30 118
2021 162 21 60
    }\plotdata
    \begin{tikzpicture}
	\begin{axis}[
	    ybar stacked,
	    /pgf/bar width=2pt,
		xlabel=Year,
		ylabel=Paper Count,
		xtick={1980, 2000, 2020},
		xticklabels={1980, 2000, 2020},
		legend style={at={(0.02,.98)},anchor=north west},
		height=2in, width=3in]
    \addplot [fill=green!70, draw=none] table [y=dialog, meta=year] {\plotdata};
    \addlegendentry{dialog(ue)}
    \addplot [fill=blue!80, draw=none] table [y=chat, meta=year] {\plotdata};
    \addlegendentry{chat}
    \addplot [fill=red!40, draw=none] table [y=conversation, meta=year] {\plotdata};
    \addlegendentry{conversation(al)}
	\end{axis}
\end{tikzpicture}
    \caption{Statistics of papers in the ACL Anthology\footnotemark{} that mention ``dialog(ue)'', ``chat'', or ``conversation(al)'' explicitly in the title over the past five decades (as of August 2021).
    This demonstrates a renewed interested in ConvAI from the ACL community since the turn of the millennium, followed by a sharp growth in the past five years.}
    \label{fig:anthology}
\end{figure}
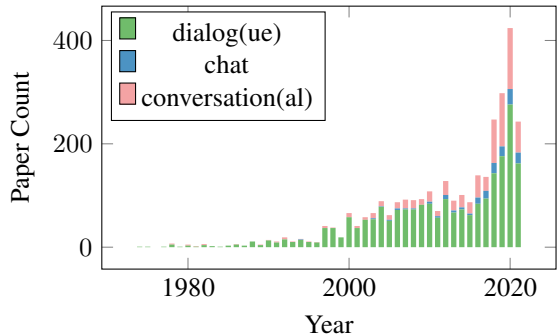

Conversational artificial intelligence (ConvAI) systems, or dialogue systems, are increasingly prevalent in our everyday lives, taking the form of smart assistants for home and hand-held devices (\eg, Amazon's Alexa, Apple's Siri, Google Assistant) or friendly helpers with telephone banking or online customer service, to name a few.
Contemporaneously, there has been a renewed interest in ConvAI from the research community, evidenced by the enthusiasm around the ConvAI challenge\footnote{\url{https://convai.io/}} and the growing number of publications in the ACL Anthology that mention ConvAI-themed keywords in their titles (see Figure \ref{fig:anthology}).

\footnotetext{\url{https://www.aclweb.org/anthology/}}

Aside from the recent technical advances, we recognize two important reasons for their popularity.
First, ConvAI systems provide a natural language user interface (LUI) for their underlying applications, requiring little to no prior training on the users' part to use them.
In an ideal world, ConvAI technology would help us build LUIs that allow users to convey their needs as easily as they would with other people.
Second, with the help of increasingly robust automatic speech recognition (ASR) and speech synthesis (or text-to-speech, TTS) systems, ConvAI systems are a lot easier and inexpensive to access than many other technical solutions.
One could gain access to a ConvAI system as long as they can access a telephone, without needing Internet access, smart devices, or even a digital cellular network.

Both reasons also make ConvAI a great candidate technology for social good, because these properties help the underlying technical solution reach a much broader population at virtually no additional cost to the society, nor delay in implementation to the communities to be served.
However, despite these inherent advantages and the growing interest in ConvAI, there has been little existing work that discusses how these systems and technologies can be developed and used for social good to the best of our knowledge.

In this paper, we aim to better understand how conversational AI technologies and systems can be developed and deployed for social good.
To this end, we begin by briefly reviewing the community's efforts and progress on ConvAI over the past few decades.
We then explore how existing technolgy can be, or have been, applied to various scenarios to advance the United Nations' {\em Sustainable Development Goals} ({\bf SDGs}) %
 for social good.
We illustrate specific use cases with concrete examples, with an focus on application scenarios that are far from common knowledge.
Finally, we conclude with reflections on potential challenges and risks when we develop and deploy new ConvAI technologies into the real world, focusing on the societal impact and how one might mitigate them.

\section{Background on Conversational AI} \label{sec:background}

Communicating with computers in human languages has been a long-standing goal of computer science since its early days.
From intuitively named commands to more recent conversational AI systems, effective use of computer systems has been made more and more accessible to more users.

For our discussion, we adopt an deliberately broad definition of the term ``Conversational AI'': \emph{any system or technology that allows human users to interact with computers via natural language}.
This definition encompasses most language-related interactive systems, including those based on audible speech interactions \citep[\eg, Carnegie Mellon University's \emph{Let's Go!} system for bus information;][]{raux2005let}.

Many early ConvAI systems give one of the conversationalists the exclusive \emph{agency} to drive the conversation.
\defn{System-initiative} ConvAI agents typically follow %
a predefined dialogue plan and offer users limited options.
As a result, these systems tend to have an easier time understanding the user as long as they are cooperating
(\eg, automated banking services).
On the other hand, \defn{user-initiative} systems %
focus on responding or reacting to user requests.
These systems can perform specific tasks at the user's request (\eg, \citepos{winograd1971procedures} SHRDLU, which manipulates geometric shapes in a block world following natural language instructions in a dialogue, or question answering systems) or keep the user company and soothed \cite{Folk_2021}.
Recent work has placed more emphasis on \defn{mixed-initiative} dialogues, where interlocutors take turns to direct the conversation.
This is also reflected in many ConvAI systems that serve a large number of users, such as XiaoIce \citep{zhou2020design} and Alexa Prize socialbots \citep{gabriel2020alexa}.

The community has also witnessed an evolution in  \emph{format} of the conversations systems are prepared to engage.
\defn{Task-oriented} conversations remain a popular problem to tackle, as they help human users to achieve concrete goals, such as booking a restaurant or turning on the light (\eg, conversations in the \multiwoz{} dataset \citep{budzianowski-etal-2018-multiwoz}).
Similar are \defn{experiment-based} dialogues where users perform experiments within an environment with the help of an ConvAI system (\eg, SHRDLU or data analytics LUIs).
These conversations often take place in a relatively closed world and involve well-framed problems.
On the other end of the spectrum, \defn{chitchat} ConvAI systems (or chatbots) engage and entertain the user without a predefined agenda, goal, or even limit on topics.
More recent research has also explored practical conversations across the spectrum of different goal specificity, domain openness, as well as levels of grounding.
Some take the familiar form of a ConvAI system assisting human users (\eg, \defn{knowledge-based} ConvAI, including question answering systems \cite{choi-etal-2018-quac}), while others focus more on conversational skills that featured in collaborative or competitive \defn{peer conversations} \citep{he-etal-2017-learning, he-etal-2018-decoupling}.

Before we move on to exploring how ConvAI systems can be applied for social good, we must first answer the following questions:
once we have built a ConvAI system, what \emph{effects} can we expect it to have on the world around us?
What \emph{properties} make them more appealing to, say, their human counterparts?
One of the most basic properties that makes ConvAI systems potentially more appealing to human operators is that they are \textbf{more readily available} and \textbf{scalable}.
This makes it possible for ConvAI services to be accessible outside typical working hours, and allow them to cater to more users at the same time at virtually no marginal cost.
As a result, ConvAI systems can be perfect alternatives to help us gather information from, or disseminate information to, a large population for social benefit with minimal requirements to the technological infrastructure.
What is more, ConvAI systems can potentially be \textbf{more easily personalized} than human operators in the future when serving a large population (e.g., adjusting the volume and/or rate of speech if the user has difficulty in hearing, or remembering user profile with consent for a more personal-level connection).
With data properly handled, they can also be better at preserving the \textbf{privacy} of individuals using them.
This makes them desirable alternatives to human interlocutors especially when the risks of personal repercussions or social stigma are perceived by the user.

\section{Opportunities for Conversational AI Systems for Social Good}

\begin{figure*}
    \centering
    \subfigure[Pandemic survey over the phone\label{fig:phone_survey:a}]{
    \resizebox{.49\textwidth}{!}{
    \parbox{0.6\textwidth}{
    \begin{framed}
    \speaker{system}: Hello, this is the CDC calling to conduct a short survey in an effort to monitor the spread of COVID-19 in the city. Is this Ms. X?\\
    \speaker{user}: \hl{Umm... Yes?}\\
    \speaker{system}: Hi Ms. X, may I ask if you have recently experienced flu-like symptoms like fever, dry cough, shortness of breath, or fatigue? \\
    \speaker{user}: \hl{No, none of those.} \\
    \speaker{system}: I see. Have you had in-person contact with anyone who recently traveled to City Y in the past month? \\
    \speaker{user}: \hl{No, not that I'm aware of. Not in the past month.}\\
    \speaker{system}: I see. Given recent developments of the pandemic, please continue to avoid indoor or outdoor gatherings, avoid unnecessary travels, and wear a mask when you are in public spaces. Thank you for your time!
    \end{framed}
    }
    }
    }
    \subfigure[Pandemic survey with emotional support
    \label{fig:phone_survey:b}]{
    \resizebox{0.45\textwidth}{!}{
    \parbox{0.6\textwidth}{
    \begin{framed}
    \begin{center}
        \textit{(system inquires about flu-like symptoms)}
    \end{center}
    \speaker{user}: \hl{Yes, I am coughing a little these two days. I'm a little concerned.}\\
    \speaker{system}: {\color{blue} \ul{I totally understand how you must feel}}. Are you experiencing fevers over 38 degrees Celsius or 100 degrees Fahrenheit?\\
    \speaker{user}: \hl{No, no fevers. Am I going to be okay?} \\
    \speaker{system}: {\color{blue} \ul{I see, no need to worry just yet!}} I have recorded your information and will flag a health worker to check-in with you later in the day. In the meantime, \emph{\color{green}please take your body temperature twice a day, drink enough fluids, and get plentiful rest. If you still feel unwell and experience fevers or chills, please visit a designated fever clinic.}\\
    \speaker{user}: \hl{Thank you, that's helpful. What other symptoms should I be watch for?}\\
    \speaker{system}: \emph{\color{green}Common symptoms also include loss of taste and smell...}
    \end{framed}
    }
    }
    }
    \caption{Two illustrative examples of pandemic survey over the phone using a ConvAI system.
    Besides gathering information for the survey (a), the system can be equipped with relatively simple natural language understanding and/or emotion detection tools to help offer {\color{blue} \ul{appeasing messages}} and \emph{\color{green}helpful advice and information} (b).}
    \label{fig:phone_survey}
\end{figure*}

In this section, we will focus on discussing and illustrating how \emph{\ul{existing}} ConvAI technologies and systems can be applied for social good.
This section is organized around how ConvAI systems can be applied to help us approach the UN's {\em Sustainable Development Goals} ({\bf SDGs}). %
For each SDG, we will focus on providing %
examples that are salient but sometimes lesser-known to illustrate how today's ConvAI technology can be applied. However we acknowledge that this is far from a comprehensive review.
As our effort to avoid the pitfall of technological solutionism, we will focus on the SDGs on which %
we see ConvAI technology having a more direct impact. %
For each application, we also analyze the limitations and boundaries of these ConvAI systems, and offer cautionary notes about areas of future development and potential risks when applicable.

\subsection{Good Health and Well-being (\sdg{3})} \label{sec:health}

As the world is swept by the COVID-19 pandemic and quarantine measures in response, health and mental well-being is
becoming more and more
the center issue on people's mind.
Many in the technology community wonder what we can do to help improve the situation, if anything.

\paragraph{Smart assistants can help inform the public.} The smart voice assistants in many homes is one way through which ConvAI technologies can help inform the public about practical guidelines in this global health crisis.
ConvAI agents can act as a source for answers to frequently asked questions regarding the pandemic, such as ``What are the common symptoms of COVID-19?'' and ``What are some best practices to prevent the spread of COVID-19?'' \citep{eddy2020mayo}.

\notes{This \emph{user-initiative}, \emph{knowledge-based} interface is only possible when we are confident the scope of questions can be relatively restricted, and can be answered from trusted sources.
That being said, it remains a technical challenge for current QA systems to know when to abstain when questions are out of scope.
Thus, such systems should always consider redirecting users to healthcare providers for further assistance and more accurate information.}

\paragraph{More effective and equitable public health measures via telephone.}
Effective and equitable public health measures should be able to reach a much broader population than just those who have access to stable Internet connections and smart devices.
Telephone surveys are an irreplaceable means for understanding the spread and effect of the pandemic \citep{worldbank2021phone}.
A timely update on a critical mass in the population is crucial to the control and monitoring of a rapidly developing pandemic, but it is also tedious and labor-intensive.
One potential way that ConvAI systems can help improve this (and similar future) situation is conducting automated surveys over the phone.
Since surveys are usually designed to be well-structured with a relatively one-sided flow of information, a ConvAI system can engage in \emph{system-initiative} conversations with a large number of users and collect their responses relatively easily (see Figure \ref{fig:phone_survey:a} for an example).
With the help of the ASR technologies, such systems can further be made to interact via natural language (instead of keypad), further making it accessible to a broader demographic to collect answers.

\notes{While accurate ASR systems require a significant amount of computational resources to serve, it is relatively easy to transmit the speech signal for centralized processing.
However, to truly realize the democratizing potentials of telephone access, ASR systems do need to be robust to low-bandwidth audio, noises and disruptions caused by unreliable cellular coverage, and underrepresented accents.}

\paragraph{Public health is not just about physical health.}
Aside from physical health, mental well-being is also of great importance, though public awareness, discovery, and intervention are still lagging behind \citep{who2013comprehensive}.
With the right set of tools, ConvAI systems have the great potential of helping us uncover potential mental issues (stress, anxiety, depression, among others), and even administer the proper intervention.
As the pandemic exerts great mental stress on the public \citep{who2020covidmental}, one need not look far from our COVID-19 survey example to find a potential candidate (see Figure \ref{fig:phone_survey:b}).

\notes{Combining the capabilities in the previous two examples, a \emph{mixed-initiative} conversation can help build rapport by actively listening to and addressing the user's concerns.
However, this does result in a significantly larger set of possible system states, and therefore require careful auditing before deployment.}

\paragraph{Safeguard mental health via early detection.}
Many people's lives have been affected by mental health issues even without the pandemic.
For instance, suicide was identified as the second leading cause of death in 2016 for young adults between 15 and 29 years of age \citep{who2019suicide}.
While chatbots have been developed for counseling and/or remedying certain mental issues \citep{han-etal-2013-counseling, fadhil-aburaed-2019-ollobot, simonite2020therapist}, they require active participants to make a difference.
On the other hand, as ConvAI systems are already widely deployed to offer customers help in \emph{task-oriented} dialogues on online platforms like e-commerce marketplaces.
With the wide reach of these online platforms, equipping ConvAI systems with a relatively simple detector of suicidal tendencies can help with early intervention before precious lives are lost (\eg, \citep{xinhua2020savelife} where one such system notified help in time to prevent a suicide attempt via drug purchase).

\notes{This is crucial for widely deployed ConvAI systems to consider, namely, what worst-case scenarios these automated systems can facilitate just by doing the exact job they are designed to do.
We should also note that, this is an example where the effect of ConvAI systems is not reflected directly in contributing to the dialogue.
Instead of engaging mentally at-risk users directly, ConvAI systems should always flag professionals for further help.}

\subsection{Quality Education (\sdg{4})}

\paragraph{Help instructors be more effective.}
Making effective use of computers to aid human instructors better achieve their educational goals has been the pursuit of the computer-assisted instruction (CAI) community for decades \cite{GRAESSER199935, yacef2002intelligentta, olney2012guru}.
Combined with domain knowledge and teaching strategies inspired by human instructors, these conversational agents can potentially distribute the experience of quality one-on-one tutoring more broadly.
This can be especially helpful in communities that experience a shortage in supply of qualified educators due to economic and/or geographic reasons, or where in-person learning is limited and engaging learners via one-to-one instruction becomes more difficult \citep{BECKER2020769}.
While teaching dialogues are often \emph{mixed-initiative}, which are more challenging, we could pursue relatively closed subsets of the conversation space by offering \emph{experiment-based} help for solving specific multi-step problems, for instance.
With direct contact to the learners, ConvAI systems can also be equipped to detect learning disabilities \citep{haavik2018conversational}.
This unique advantage will not only allow the ConvAI systems themselves to adapt, but also potentially inform the human instructors to better cater to each learner's unique learning needs.

\notes{In many cases, ConvAI needs to work together with other forms of user interface to provide an effective extension to the learning experience, \eg, real or simulated physical or chemical experiments to help students explore in physics and chemistry.
We also note that the diagnosis and intervention of learning disabilities should be left to education professionals, with ConvAI in a role to aid early flagging or approved assistance in intervention.}

\paragraph{Available beyond the classroom.}
Besides improving the bandwidth and teaching style of traditional classroom learning, ConvAI systems can also engage learners better outside the physical or virtual classroom.
Having exclusive access to a virtual learning assistant can potentially help eliminate the effect of imbalance in the allocation of limited teaching resources (\eg, asking questions in class).
Not being bound to limited office hours, ConvAI systems can act as a bridge to help alleviate the pressure of non-anonymity, peer pressure, or time pressure in teacher-student interactions.
Moreover, these agents can be personalized to fit each learner's habit and better help each individual tackle long-term goals, such as building one's vocabulary or preparing for a test, with helpful reminders, check-ins, and interactive exercises.

\notes{While some of the functionality mentioned above do not necessitate them, ConvAI systems could potentially provide an intuitive, unified, and interactive language user interface to lower the barrier to make effective use of them.}

\subsection{Reducing Inequalities (\sdg{10})}

Our society evolves at breakneck speed, especially when one retrospects on the technological advances over the recent decades.
In the meantime, social inequalities in opportunities also increase as different countries, communities, or individuals benefit from these advances differently.
ConvAI is well-positioned to reduce some of these inequalities.

\paragraph{Equitable policies begin with equitable access to governments.}
One of the essential needs that ConvAI systems are well-equipped to help address is that each community can easily express their needs to local officials to inform policy-making.
Similar to the telephone survey we described in Section \ref{sec:health}, a \emph{system-initiative} ConvAI agent over the phone can help gather community feedback while preserving the privacy or identity of callers \cite{ANDROUTSOPOULOU2019358}.
This allows users to share their opinions on poignant issues (\eg, housing, public safety) freely without worrying about stigma or repercussions.
ASR transcripts can further be triaged and clustered with NLP tools for government officials to process more efficiently, further expanding the bandwidth in public opinion intake.

\notes{While privacy preservation can be achieved technically, stakeholders should operate with transparency to ensure it is known to and trusted by the public to achieve the desired outcome.
Abuse under anonymity could also be a potential risk to mitigate.}

\paragraph{Disability and quality of life should not be mutually exclusive.}
Aside from accessing the public discourse, it is also important that everyone is able to enjoy their private life, especially the benefits brought by technological development.
People with disabilities are too often left behind by the convenience of modern life.
NLP has great potential in improving life quality for the visually and hearing impaired through assistive technologies (\eg, audio description for movies \citep{rohrbach2017movie} and closed captioning via ASR).
However, many existing assistive language technologies provide one-sided interfaces where people are only receiving information from the system.
We argue that ConvAI technologies can be applied in many of these applications to bring an interactive experience that further extend the potential for them.
For instance, augmented with \emph{knowledge-based} visual question answering \citep{antol2015vqa} and visual dialogue \citep{das2017visual}, movie description systems can further help the visually impaired explore scenes in creative art, providing a fuller viewing experience.

\notes{Despite the technical plausibility of interactive assistive technology, it is nevertheless essential to involve potential users to understand whether/how they desired the technology to take shape in innovations like this.}

\paragraph{Fight inequality by reinforcing existing mechanisms.}
Inequality is not new to our society, and it predates most of the technological advances we have discussed in this paper.
To reduce inequality brought about by various societal changes, we have collectively come up with many solutions in human history, one of which is charity organizations that help redistribute resources to people and causes in need.
Besides directly helping people affected by inequality, can ConvAI technologies benefit social good by making a positive impact on these organizations?
Recently, \citet{wang-etal-2019-persuasion} demonstrated an inspiring angle to answer this question, where ConvAI agents are trained to persuade the interlocutor to make charitable donations to certain causes.
They showed that these agents can potentially be personalized to the psychological backgrounds of different persuadees to promote positive outcome.
For instance, when attempting to persuade someone who is more open (as defined in the Big-Five traits \cite{goldberg1992development}), inviting them to learn about the cause/organization (termed \emph{source-related inquiry} by \citet{wang-etal-2019-persuasion}) could be helpful, \eg, ``Have you heard of Save the Children?'' or ``Are you familiar with the organization?''.

\notes{One salient risk with persuasive technology is the use to persuade people to do harm, or for economic or political advantages.
We urge the use of these technologies be regulated, and required to disclose the motivation, target, beneficiary, etc.~for transparency.}

\subsection{Shared Economic Growth (\sdgs{1}{8})}

Social inequalities within and between countries are often rooted in the socioeconomic statuses of the population.
It is imperative that we consider technological solution to not only reduce existing inequalities, but also eradicate its source when possible.
Perhaps one of the most important and sustainable approach towards ending poverty is providing decent jobs with reasonable pay, where the oft-neglected process of data collection can potentially help.

\paragraph{ConvAI can positively affect people aside from directly serving them.}
Despite our focus so far on applications of ConvAI for social good, one aspect that is commonly neglected is how these systems are built.
Data is critical to modern (Conv)AI systems, and the process of acquiring this data often involve paid annotation teams working in a controlled setting.
The nature of ConvAI makes it a great candidate for redistributing the payout of data collection to the disadvantaged, because: (1) natural conversation data is in great need, which is easy to generate without too much training  \cite[\eg, using a Wizard-of-Oz approach, see ][]{kelley1984iterative, budzianowski-etal-2018-multiwoz}, and can usually be collected in a safe environment;\footnote{Here, we focus on the raw speech/text data, and acknowledge that they usually require separate post-processing.} (2) ConvAI data can be delivered via telephone/text messaging (SMS) systems if necessary, which is much more widely available; (3) as a result of the wider reach, the resulting data is also more diverse, and will help ConvAI systems better adapt to different data variations, including those that would naturally occur in some of the aforementioned scenarios where ConvAI can be applied (over the phone or SMS).

\notes{Data collection should only be conducted with full knowledge and consent from participants, with privacy protection measures where applicable.
Given the prevalent income inequality, good pay in economically disadvantaged countries/regions can be economic for affluent ones, where researchers and practitioners are typically based.
This is no excuse for unethical, exploitative low pay, however \citep{gray2019ghost}.}

\section{Challenges for Conversational AI Systems for Social Good} \label{sec:challenges}

In the previous section, we have explored various opportunities that current ConvAI technologies can contribute to social good.
In this section, we turn to potential challenges we need to tackle to better realize this goal, and articulate some of the salient risks that might lie within the further development and deployment of ConvAI technology in the real world.

\subsection{Shared Challenges with ML/NLP for Social Good}

Before diving into issues that are more unique to ConvAI, we briefly review some of the key challenges and considerations that CovnAI systems share with other machine learning (ML) or natural language processing (NLP) technologies when applied for social good.

\paragraph{Problem-centric, not tech-centric.} One of the common pitfalls is technology-centered solutionism.
\citet{tuomi2018jrc} summarized its origins and risks aptly in an European Union Science for Policy report on AI's impact on education:
\begin{displayquote}
\emph{In the stage of technology push, technology experts possess scarce knowledge
... [which]
often dominates and overrides other types of knowledge ...
this can become a problem as technologists easily transfer their own experiences and beliefs about learning to their designs.}
\end{displayquote}

Although technologists do bring fresh perspectives, it is crucial to remember that when developing technology to solve problems, domain experts are more knowledgeable of the mechanisms, causes, and nuances regarding the subject matter.
\citet{green2019good} also warns us against the dangers of underdefined or short-sighted metrics of social good, which are sometimes the result of lack of communication between technologists and subject matter experts.

By extension, it is also important to assess whether ConvAI (or other approaches familiar to the technologists) is significantly better than alternative investments.
While requiring simpler equipment to deliver, language interfaces are not always the most efficient if graphical user interfaces are applicable (\eg, providing directions in a building).

\paragraph{Avoid Amplifying Bias.} It is also important to avoid exacerbating existing social biases or inequalities.
Today, this is commonly rooted in the kind of data available for many ML applications, from facial recognition \citep{garvie2016facial} to automatic speech recognition \citep{koenecke2020racial}, to what is more and more recognized in virtually any NLP application, the lack of linguistic diversity \citep{joshi-etal-2020-state}.
It is important to carefully design and curate the data used to build these systems, especially in a social good setting to avoid making a bad situation worse.

\subsection{Technical challenges for ConvAI}

In this section, we will focus on technical challenges and research frontiers that are more closely (but not necessarily exclusively) related to ConvAI, and share our view on the path forward, and the main obstacles therein.

\paragraph{Know thy interlocutor.} One of the fundamental goals of ConvAI is to facilitate communication, therefore it is important to take into consideration whether the system is communicating in a manner that is clear and effective for the interlocutor.
This is particularly important considering the increased diversity of population we set out to reach in applications for social good.
One aspect to factor into consideration when designing ConvAI systems to communicate with people from diverse backgrounds is to understand how they would perceive the same message or mode of communication, as miscommunication can sometimes easily slip detection and lead to catastrophic outcomes \citep{wiki:avianca52}.
We urge researchers to consider using guidelines like Datasheets \cite{gebru2018datasheets} to help calibrate the collection and use of data, and when possible, involve members of communities the system is intended for in the process.

\paragraph{Go beyond words.} Besides a better understanding of users that the system is designed to interact with, we can also make ConvAI systems communicate more effectively with the help of other modalities than just text and speech.
As interactive technologies like high-quality computer graphics, robust computer vision, haptics devices, augmented and virtual reality (AR/VR), and embodied technology become increasingly available to the public, there are also great opportunities for computer scientists from these different areas to join forces in the quest towards better interactive computer systems \citep[\eg,][]{baai2020jddc}.
As we have discussed, one salient application that will benefit from multimodal interactions is education.
These technologies will help broaden our horizon, and potentially help enable interactive conversational systems for social good that is beyond our current imagination.
However, some of the obstacles of further developments in this area, we believe, include data sharing and interdisciplinary collaboration.

\paragraph{Faster, Higher, Greener.} As we pack more features into ConvAI systems, it is always useful to remember that at the end of the day, these systems are built to interact with humans in real time.
Thus, besides being able to communicate effectively, reasonable response time is of great importance compared to off-line ML systems.
On the other hand, the great opportunity for ConvAI to help in various settings and grounded situations should also encourage us to investigate better methods for transferring between different settings with higher data-efficiency.
For instance, once we have built a well-designed system for pandemic survey for COVID-19, it can ideally help with surveying other infectious diseases without significant effort in data collection, training, or system configuration.
Both of these goals are not only practical and desirable for ConvAI developers and social good stakeholders, but potentially also helpful in reducing the computational cost and climate impact of these systems while they are serving to address societal problems (\sdgstext{12 Responsible Consumption and Production}{13 Climate Action}).
We believe that the rise of multi-domain dialogue datasets \citep[\eg, \multiwoz, ][]{budzianowski-etal-2018-multiwoz} are a great first step in this area.

\paragraph{It takes two... or more.} Although so far in this paper we have largely restricted our attention to conversations between one human user and one ConvAI system, this is by no means the only form of conversation that naturally occurs, or in the context of social good.
One of the common scenarios that multi-party conversations naturally occur is peer support groups (\sdg{3}), where ConvAI can potentially serve as the coordinator between multiple human speakers to help guide the conversation \citep{nordberg2019designing}.
With the growing popularity of online forums, automatic moderation of online discussions has also received ample research attention in recent years \citep{delort2011automatic}.
We believe there is still a large space of exploration for what multi-party ConvAI systems can help us achieve in building healthier online communities large and small.

\subsection{Social challenges for ConvAI}

Now that we have discussed the technical challenges facing ConvAI and its application in social good in the near future, we turn to social challenges that are unlikely addressed solely by technical advances, and/or that require a wider societal awareness and collective effort to mitigate.

\paragraph{Data sourcing and quality.}
The source of training and evaluation data is one of the most important aspects to consider when building and deploying modern AI systems.
This is because not only does data affect the behavior of these systems, but it also often functions as a standard to compare and evaluate different candidates before ``better'' systems are chosen and deployed.

As we have discussed in our section on shared challenges, it is important that we don't exacerbate existing social biases when building these systems, and one of the most fundamental ways to alleviate this bias is to collect data that is representative of the target audience of our system. (\sdg{10})
Aside from representation of the population, what can usually be neglected is the representation of actual users of the system.
In the ConvAI research literature, many datasets are collected from paid crowdworkers rather than actual users that the systems are intended to serve, which might not faithfully represent user behavior \citep{devries2020towards}.
Furthermore, as most datasets are collected via a Wizard-of-Oz approach, even if we were able to simulate the intended users with crowdworkers, they will be conversing with a ``perfect'' system that doesn't make certain mistakes that a trained ConvAI system would.
This divide is crucial to realize and mitigate in future work, especially for real-world applications.

Another challenge that threatens the data quality for ConvAI concerns systems that adapt on the fly to data collected during the time they are serving.
On one hand, overly relying on usage data could potentially obfuscate or even exacerbate the impact of blind spots in the system, as humans tend to adjust their language use to accommodate the perceived properties of the interlocutor \citep{giles1991accommodating}.
If a component in a ConvAI system does not meet a user's initial expectation, instead of trying the same thing over and over again, they are much more likely to accommodate the system by either enunciating or avoid the component entirely if possible.
It is thus important to look beyond the survivorship bias that is usually present in such usage data.
On the other hand, systems made publicly available are more vulnerable to data poisoning attacks or adversarial data manipulation \citep{nelson2008exploiting, rubinstein2009antidote}.
Without sufficient defense and audit mechanisms, a well-intended system can easily and quickly be polluted through its vulnerabilities, and turned against the very people it set out to serve \cite[\eg, Microsoft's Tay bot, which was misled by unseen, unhandled malicious data;][]{vincent2016tay}.

\paragraph{Data sharing.}
For the sustained development of ConvAI  and its adoption in social good, a coordinated effort between academia and the industry is also essential.
Despite the abundance of talent and research freedom, academia usually lacks the means to obtain substantial amounts of data for ConvAI, especially data that reflect the ``realistic data distribution''.
In contrast, the industry has a broader access to real-world data and problems, yet usually limited research staff to tackle all of them.
While their interests somewhat diverge between academically interesting, abstract, and open research projects (academia) versus proprietary technology and practical solutions for problems at hand (industry), we would like to highlight that their shared common interests in empirically sound technology that hopefully satisfy a wide range of real-world needs.
Therefore, we would like to advocate closer collaborations between the two on how to collect, convert, anonymize, and share conversational data (especially data less proprietary in nature) and work on repeatable evaluation methods, so that the broader research community can be exposed to unique problems rooted in real-world issues that can generate great positive impact.
It should be stressed, however, that user privacy and consent are integral to any effort to share user-related data.

\paragraph{Anthropomorphism and aggression from human users.}
As ConvAI systems are deployed in the wild, they are commonly perceived by users as having human-like traits or characteristics, which is especially salient for video-/voice-based systems.
In the meantime, it is not uncommon that they meet aggression from human users from time to time, and managing user frustration is an important topic.
Beyond helping the user achieve their goals, however, we should also watch out for how these interactions can potentially affect our interaction with other humans.

Most ConvAI systems available today on the market project a female persona (exclusively or by default), which is even more pronounced when they interact using speech \citep{tay2014stereotypes, unesco2019blush}.
As they inevitably fail to fulfill our requests from time to time, these systems can become the target of frustration, which may also inadvertently affect how the new generation interact with ConvAI systems as they grow up around them \cite{elgan2018kidsalexa, rudgard2018alexachildren}.
Setting aside the debate about child education, it is important and responsible that we better understand whether this phenomenon mirrors actual human interactions, and how it would contribute to social interactions in the long term.
If these devices are helping instill or reinforce stereotypes of genders and gender dynamics, it is the responsibility of the community to raise public awareness and help advocate means to mitigate this effect (\sdg{5 Gender equality}).

\paragraph{Misuse for Deception.}
Technological advances are almost always double-edged swords.
As NLP systems become more advanced, it will inevitably be easier to make them appear more human-like.
For instance, GPT-3 \citep{brown2020language} was used to post on behalf of an account on Reddit, and because of its coherent and convincing posts, it went undetected for weeks \citep{heaven2020gpt3reddit}.

This poses a higher risk in ConvAI systems due to their interactive nature, as their misuse can lead to more elaborate and convincing deception schemes such as spreading mis/disinformation or scamming.
Seemingly benign technology that allow ConvAI systems to assume a coherent identity \citep{li-etal-2016-persona} can potentially help someone impersonate someone else online, akin to how generative adversarial networks \citep{goodfellow2014generative} have lead to convincing deepfakes.
This would not only jeopardize causes for social good that we have advocated for in this paper, but can potentially cause significant damage to our society if left unchecked.
We urge the community to reflect on these potential damages, and work with governments to regulate their use in legit businesses, as well as educate the public about their risks.

\paragraph{Transparency and trustworthiness.}
When ConvAI systems interact with the general population, they face the choice of whether to declare their identity as a computer system or robot.
This active act of transparency is not always designed into systems for various reasons \citep[\eg, Google Duplex when it was initially revealed;][]{google2018duplex}, and as a result might sometimes have unintended consequences \citep{garun2019duplexrestaurants} despite Duplex's efforts to self-identify \citep{fagan2018google}.
Although the lack of transparency in this case is not a deceptive act for malignant purposes, the restaurants' perplexed reaction to Duplex clearly indicates that it is not inconsequential.

Another aspect related to transparency is trustworthiness, which is essential for any AI system, especially ConvAI systems that interact with humans directly.
To this end, we argue that ConvAI systems deployed in the real world should be able to recognize their limits of capabilities, and communicate these limits clearly.
This can be a crucial step in earning trust especially in the event that the system has misstepped as a result of misidentifying user intent and sentiment, or even mischaracterized user profile in an attempt to personalize to the user.

\section{Concluding Remarks}

In this paper, we have introduced how conversational artificial intelligence (ConvAI) systems can be applied to causes for social good today and in the near future, and summarized some of the major technical and social challenges the community still needs to tackle down the line.
We acknowledge that the opportunities and risks we included in this paper are by no means comprehensive, but an attempt of us to summarize and introduce what in our opinion are typical and atypical scenarios where ConvAI can be impactful or concerning to the community.
We refer the reader to \citep{ruane2019conversational} for a more comprehensive view of social and ethical considerations of ConvAI.

This paper is part of a broader conversation around the impact of AI systems in the real world, including the very definition of social good \citep{green2019good}, transparency and interpretablility \citep{doshi2017towards}, as well as algorithmic fairness \citep{corbett2017algorithmic}, among others.
We hope that our paper can serve as a starting point for the community brainstorming of various opportunities and potential risks for ConvAI to promote social good and betterment.

{\small
\bibliography{anthology,main}
}

\end{document}